\RequirePackage{fix-cm}
\documentclass[twocolumn]{svjour3}          
\smartqed  
\usepackage{xcolor}

\usepackage{mathptmx} 

\usepackage{amsmath} 
\usepackage{amssymb}  

\usepackage{bm}
\usepackage[short]{optidef}
\usepackage{algorithm}
\usepackage[noend]{algpseudocode}

\usepackage{tikzscale}
\usepackage{pgfplots}
\pgfplotsset{compat=1.9, legend style={font=\footnotesize}}
\usepgfplotslibrary{groupplots}
\usepackage{setspace}
\newcommand{\hl}[1]{{ #1}} 

\usepackage{hyperref}
\hypersetup{
    colorlinks=true,
    linkcolor=blue,
    filecolor=magenta,      
    urlcolor=black,
}
\usepackage{booktabs}

\newcommand{\specialcellbold}[2][c]{%
	\bfseries
	\begin{tabular}[#1]{@{}l@{}}#2\end{tabular}%
}

\journalname{Autonomous Robots}
\begin{document}

\title{ALGAMES: A Fast Augmented Lagrangian Solver for Constrained Dynamic Games
}


\author{Simon Le Cleac'h        \and
        Mac Schwager             \and
        Zachary Manchester 
}



\institute{
    Simon Le Cleac'h \at
    Department of Mechanical Engineering, 
    Stanford University, USA \at
    simonlc@stanford.edu        
\and
    Mac Schwager \at
    Department of Aeronautics \& Astronautics,
    Stanford University, USA \at
    schwager@stanford.edu        
\and 
    Zachary Manchester \at
    The Robotics Institute, 
    Carnegie Mellon University, USA \at
    zacm@cmu.edu          
}

\date{Received: date / Accepted: date}

\maketitle 

\begin{abstract}
Dynamic games are an effective paradigm for dealing with the control of multiple interacting actors. This paper introduces ALGAMES (Augmented Lagrangian GAME-theoretic Solver), a solver that handles trajectory-optimization problems with multiple actors and general nonlinear state and input constraints. Its novelty resides in satisfying the first-order optimality conditions with a quasi-Newton root-finding algorithm and rigorously enforcing constraints using an augmented Lagrangian method. We evaluate our solver in the context of autonomous driving on scenarios with a strong level of interactions between the vehicles. We assess the robustness of the solver using Monte Carlo simulations. It is able to reliably solve complex problems like ramp merging with three vehicles three times faster than a state-of-the-art DDP-based approach. A model-predictive control (MPC) implementation of the algorithm, running at more than 60 Hz, demonstrates ALGAMES' ability to mitigate the ``frozen robot'' problem on complex autonomous driving scenarios like merging onto a crowded highway.
\vspace{-2mm}
\vspace{-2mm}
\end{abstract}

\noindent\rule{8.5cm}{0.4pt} \\
\footnotesize{}
This work was supported in part by NSF NRI grant 1830402, DARPA grant D18AP00064 and ONR grant N00014-18-1-2830.  Toyota Research Institute (``TRI'') provided funds to assist the authors with their research, but this article solely reflects the opinions and conclusions of its authors and not TRI or any other Toyota entity.
\normalsize{}

    \begin{figure}[t]
    \centering
    \includegraphics[width=8.3cm]{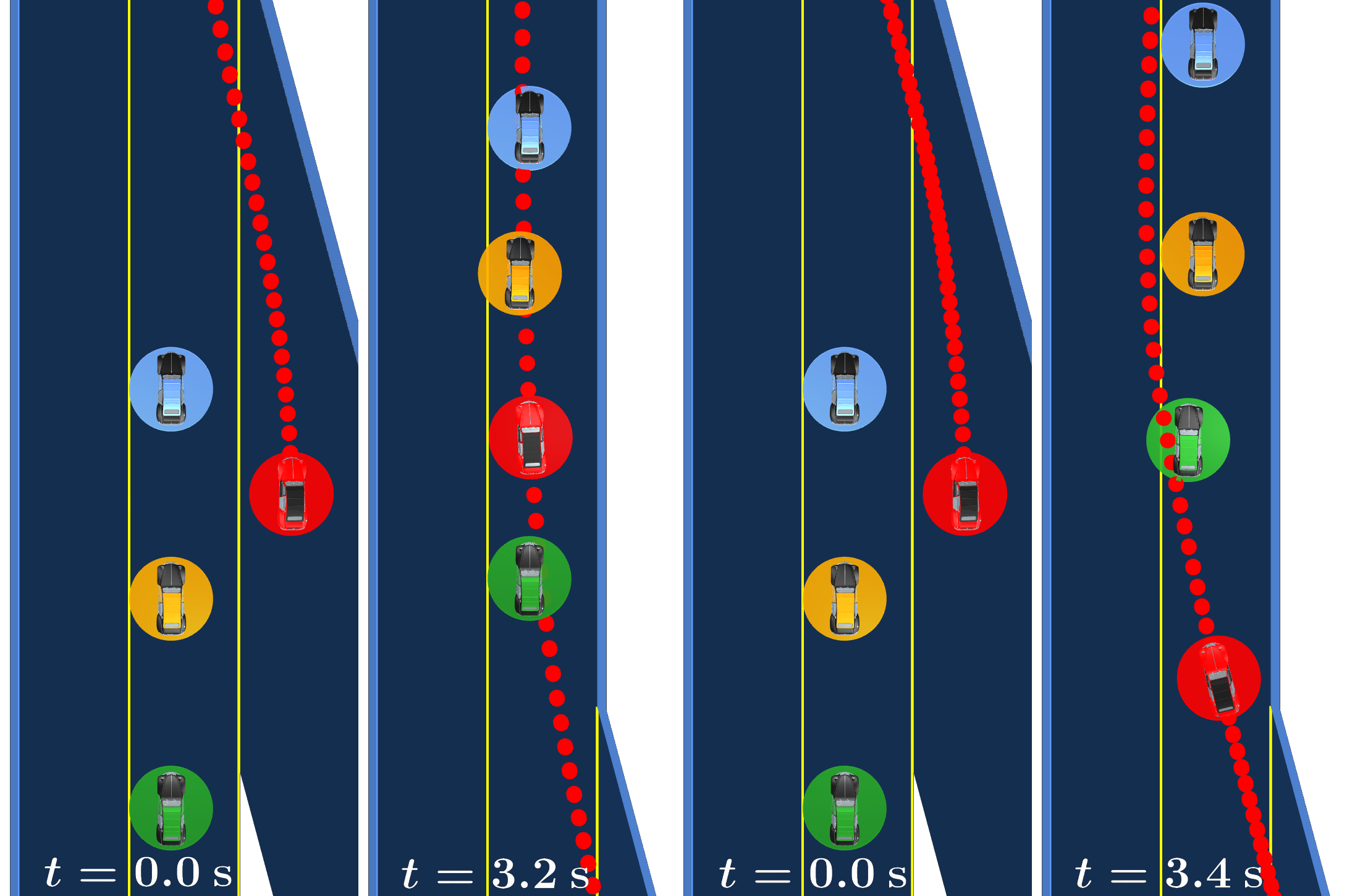}\hfill
    \caption{A merging maneuver on a crowded highway is carried out using a receding horizon implementation of ALGAMES. On the left, the red car controlled by ALGAMES merges between the orange and green cars with minimal disruption to the traffic. On the right, the red car controlled by a non-game-theoretic MPC, is ``frozen,'' i.e. it cannot find a feasible path. It has to slow down significantly and wait for the three cars to pass before merging.}
    \label{fig:mpc_frozen_robot}
    \end{figure}


\section{Introduction}
    Controlling a robot in an environment where it interacts with other agents is a complex task. Traditional approaches in the literature adopt a predict-then-plan architecture. First, predictions of other agents' trajectories are computed, then they are fed into a planner that considers them as immutable obstacles. This approach is limiting because the effect of the robot's trajectory on the other agents is ignored. Moreover, it can lead to the ``frozen robot'' problem that arises when the planner finds that all paths to the goal are unsafe \cite{Trautman2010}. It is, therefore, crucial for a robot to \emph{simultaneously} predict the trajectories of other vehicles on the road while planning its own trajectory, in order to capture the reactive nature of all the agents in the scene.  ALGAMES provides such a joint trajectory predictor and planner by considering all agents as players in a Nash-style dynamic game.  We envision ALGAMES being run on-line by a robot in a receding-horizon loop, at each iteration planning a trajectory for the robot by explicitly accounting for the reactive nature of all agents in its vicinity.

    Joint trajectory prediction and planning in scenarios with multiple interacting agents is well-described by a dynamic game.  Dealing with the game-theoretic aspect of multi-agent planning problems is a critical issue that has a broad range of applications. For instance, in autonomous driving, ramp merging, lane changing, intersection crossing, and overtaking maneuvers all comprise some degree of game-theoretic interactions \cite{Sadigh2016}, \cite{Sadigh2016a}, \cite{Fridovich-Keil2020a},  \cite{Dreves2018},  \cite{Fisac2019},  \cite{Schmerling2018}. Other potential applications include mobile robots navigating in crowds, like package-delivery robots, tour guides, or domestic robots; robots interacting with people in factories, such as mobile robots or fixed-base multi-link manipulators; and competitive settings like drone and car racing \cite{Wang2021},  \cite{Liniger2019}. 
    
    In this work, we seek solutions to constrained multi-player dynamic games. In dynamic games, the players' strategies are sequences of decisions. It is important to notice that, unlike traditional optimization problems, non-cooperative ga-mes have no ``optimal'' solution. Depending on the structure of the game, asymmetry between players, etc., different concepts of solutions are possible. In this work, we search for Nash-equilibrium solutions. This type of equilibrium models symmetry between the players; All players are treated equally. At such equilibria, no player can reduce its cost by unilaterally changing its strategy. For extensive details about the game-theory concepts addressed in this paper, we refer readers to the work of Basar et al. \cite{Basar1999}.

    Our solver is aimed at finding a Nash equilibrium for multi-player dynamic games, and can handle general nonlinear state and input constraints. This is particularly important for robotic applications, where the agents often interact through their desire to avoid collisions with one another or with the environment. Such interactions are most naturally represented as (typically nonlinear) state constraints. This is a crucial feature that sets game-theoretic methods for robotics apart from game-theoretic methods in other domains, such as economics, behavioral sciences, and robust control. In these domains, the agent interactions are traditionally represented in the objective functions themselves, and these games typically have no state or input constraints. In mathematics literature, Nash equilibria with constraints are referred to as \emph{Generalized Nash Equilibria} \cite{Facchinei2007}.  Hence, in this paper we present an augmented Lagrangian solver for finding Generalized Nash Equilibria specifically tailored to robotics applications.  
    
    Our solver assumes that players are rational agents acting to minimize their costs. This rational behavior is formulated using the first-order necessary conditions for Nash equilibria, analogous to the Karush-Kuhn-Tucker (KKT) conditions in optimization. By relying on an augmented Lagrangian approach to handle constraints, the solver is able to solve multi-player games with several agents and a high level of interactions at real-time speeds. Finding a Nash equilibrium for 3 autonomous cars in a freeway merging scenario takes $90$ ms. Our primary contributions are:
    \begin{enumerate}
        \item A general solver for dynamic games aimed at identifying Generalized Nash Equilibrium strategies.
        \item A real-time MPC implementation of the solver mitigating the ``frozen robot" problem that arises in complex driving scenarios for non-game-theoretic MPC approach. (Fig. \ref{fig:mpc_frozen_robot}).
        \item An analysis of the non-uniqueness of Nash equilibria in driving scenarios with constraints and an assessment of the practical impact it has on players' coordination.
        \item A comparison with iLQGames \cite{Fridovich-Keil2020a}. ALGAMES finds Nash equilibria 3 times faster than iLQGames for a fixed constraint satisfaction criterion.
    \end{enumerate}

\section{Related Work}

    \subsection{Equilibrium Selection}
    Recent work focused on solving multi-player dynamic games can be categorized by the type of equilibrium they select. Several works \cite{Sadigh2016}, \cite{Sadigh2016a}, \cite{Liniger2019}, \cite{Yoo2012} have opted to search for Stackelberg equilibria, which model an asymmetry of information between players. These approaches are usually formulated for games with two players: a leader and a follower. The leader chooses its strategy first, then the follower selects the best response to the leader's strategy. Alternatively, a Nash equilibrium does not introduce hierarchy between players; each player's strategy is the best response to the other players' strategies. As pointed out in \cite{Fisac2019}, searching for open-loop Stackelberg equilibrium strategies can fail on simple examples. In the context of autonomous driving, for instance, when players' cost functions only depend on their own state and control trajectories, the solution becomes trivial. The leader ignores mutual collision constraints and the follower has to adapt to this strategy. This behavior can be overly aggressive for the leader (or overly passive for the follower) and does not capture the game-theoretic nature of the problem. 
    
    Nash equilibria have been investigated in \cite{Fridovich-Keil2020a}, \cite{Dreves2018}, \cite{Wang2021}, \cite{Britzelmeier2019}, \cite{Di2018}, \cite{Di2019}, \cite{Di2020a}. We also take the approach of searching for Nash equilibria, as this type of equilibrium seems better suited to symmetric, multi-robot interaction scenarios. Indeed, we have observed more natural behavior emerging from Nash equilibria compared to Stackelberg when solving for open-loop strategies.


    \subsection{Game-Theoretic Trajectory Optimization}
    Most of the algorithms proposed in the robotics literature to solve for game-theoretic equilibria can be grouped into four types: First are algorithms aimed at finding Nash equilibria that rely on decomposition, such as Jacobi or Gauss-Siedel methods \cite{Wang2021}, \cite{Britzelmeier2019}. These algorithms are based on an iterative best-response (IBR) scheme in which players take turns at improving their strategies considering the other agents' strategies as immutable \cite{Facchinei2007}. This type of approach is easy to interpret and scales reasonably well with the number of players. However, convergence of these algorithms is not well understood \cite{Facchinei2007}, and special care is required to capture the game-theoretic nature of the problem \cite{Wang2021}.  Moreover, solving for a Nash equilibrium until convergence can require many iterations, each of which is a (possibly expensive) trajectory-optimization problem. This can lead to prohibitively long solution times.

    Second, there are a variety of algorithms based on dynamic programming. In \cite{Fisac2019}, a Markovian Stackelberg strategy is computed via dynamic programming. This approach seems to capture the game-theoretic nature of autonomous driving. However, dynamic programming suffers from the curse of dimensionality and, therefore, practical implementations rely on simplified dynamics models coupled with coarse discretization of the state and input spaces. To counterbalance these approximations, a lower-level planner informed by the state values under the Markovian Stackelberg strategy is run. This approach, which scales exponentially with the state dimension, has been demonstrated in a two-player setting. Adding more players is likely to prevent real-time application of this algorithm. In contrast, our proposed approach scales polynomially with the number of players (see Section \ref{sec:complexity}).
    
    \hl{Third, algorithms akin to differential dynamic programming have been developed for robust control} \cite{Morimoto2003}, \cite{Sun2015},  \hl{and later applied to game-theoretic problems} \cite{Fridovich-Keil2020a}, \cite{Di2018}. \hl{Similar methods were applied in the stochastic} \cite{Sun2016}, \hl{and belief-space planning settings} \cite{Schwarting2021}. This approach scales polynomially with the number of players and is fast enough to run real-time in a MPC fashion \cite{Fridovich-Keil2020a}. However, contrary to ALGAMES, this type of approach does not natively handle constraints. Collision-avoidance constraints are typically handled using large penalties that can result in numerical ill-conditioning which, in turn, can impact the robustness or the convergence rate of the solver. Moreover, it leads to a trade-off between trajectory efficiency and avoiding collisions with other players.
    
    Finally, algorithms that are analogous to direct methods in trajectory optimization have also been developed \cite{Di2019}, \cite{Di2020a}. An algorithm based on a first-order splitting method that is known to have a linear convergence rate was proposed by Di et al. \cite{Di2020a}. Di's experiments show convergence of the algorithm after typically $10^3$ to $10^4$ iterations. A different approach based on Newton's method has been proposed \cite{Di2019}, but it is restricted to unconstrained dynamic games. Our solver belongs to this family of approaches. It also relies on a second-order Newton-type method, but it is able to handle general state and control input constraints. In addition, we demonstrate convergence on relatively complex problems in typically less than $10^2$ iterations. 
    
    \subsection{Generalized Nash Equilibrium Problems}
    
    We focus on finding Nash equilibria for multi-player games in which players are coupled through shared state constraints (such as collision-avoidance). Therefore, these problems are instances of Generalized Nash Equilibrium Problems (GNEPs). The operations research field has a rich literature on GNEPs \cite{Pang2005}, \cite{Facchinei2006}, \cite{Facchinei2009}, \cite{Facchinei2010}, \cite{Fukushima2011}. Exact penalty methods have been proposed to solve GNEPs \cite{Facchinei2006}, \cite{Facchinei2009}. Complex constraints such as those that couple players' strategies are handled using penalties, allowing solution of multi-player games jointly for all the players. However, these exact penalty methods require minimization of nonsmooth objective functions, which leads to slow convergence rates in practice.
    
    In the same vein, a penalty approach relying on an augmented Lagrangian formulation of the problem has been advanced by Pang et al. \cite{Pang2005}. This work, however, converts the augmented Lagrangian formulation to a set of KKT conditions, including complementarity constraints. The resulting constraint-satisfaction problem is solved with an off-the-shelf linear complementarity problem (LCP) solver that exploits the linearity of a specific problem. Our solver, in contrast, is not tailored for a specific example and can solve general GNEPs. It draws inspiration from the augmented Lagrangian formulation, which does not introduce nonsmooth terms in the objective function, enabling fast convergence. Moreover, this formulation avoids ill-conditioning, which improves the numerical robustness of our solver. 
    
    \hl{Another solution method based on KKT conditions for constrained dynamic games has been proposed} \cite{Schwarting2019b}. \hl{It consists of stacking together the individual KKT conditions of all the players and solve the resulting problem as a single optimization using an off-the-shelf interior point solver (IPOPT)} \cite{Wachter2006}. \hl{This approach has been explored with box constraints but not with more complicated inequality constraints such as collision avoidance. Indeed, when the individual problems contain inequality constraints, complementarity constraints are introduced in the stacked KKT system. Those constraints are not natively handled by interior point solvers as the KKT system has no interior at the solution. They also violate constraint qualifications that are required for well-posed, nonlinear programs} \cite{Biegler2010}. \hl{Off-the-shelf solvers typically rely on iterative relaxation schemes that may lead to solution times precluding their use in an MPC framework. ALGAMES is less general than a general-purpose optimization package such as IPOPT, but is targeted specifically at solving GNEPs involving complex inequality constraints with an augmented Lagrangian scheme that allows for fast solution times.} 
    
    
    

\section{Problem Statement}
    In the discretized trajectory-optimization setting with $N$ time steps, we denote by $n$ the state size, $m$ the control-input size, $x_k$ the state, and $u^{\nu}_k$  the control input of player $\nu$ at the time step $k$. In formulating the game, we do not distinguish between the robot carrying out the computation, and the other agents whose trajectories it is predicting.  All agents are modeled equivalently, as is typical in the case of Nash-style games.

    Following the formalism of Facchinei \cite{Facchinei2007}, we consider the GNEP with $M$ players. Each player $\nu$ decides over its control input variables $U^{\nu} = [({u_1^{\nu}})^T \ldots ({u_{N-1}^{\nu}})^T]^T \in \mathbb{R}^{\bar{m}^\nu}$. This is player $\nu$'s strategy where $m^{\nu}$ denotes the dimension of the control inputs controlled by player $\nu$ and $\bar{m}^{\nu} = m^{\nu}(N-1)$ is the dimension of the whole trajectory of player $\nu$'s control inputs. By $U^{-\nu}$, we denote the vector of all the players' strategies except the one of player $\nu$. Additionally, we define the trajectory of state variables $X = [({x_2})^T \ldots ({x_{N}})^T]^T \in \mathbb{R}^{\bar{n}}$ where $\bar{n} = n(N-1)$, which results from applying all the control inputs decided by the players to a joint dynamical system,
    \begin{align}
        x_{k+1} = f(x_k, u^1_k, \ldots, u^M_k) = f(x_k, u_k),
    \end{align}
    with $k$ denoting the time-step index. The kinodynamic constraints over the whole trajectory can be expressed with $\bar{n}$ equality constraints,
    \begin{align}
        D(X, U^1, \ldots, U^M) = D(X, U) = 0 \: \in \mathbb{R}^{\bar{n}}.
        \label{eq:dynamics}
    \end{align}
    The cost function of each player is noted $J^{\nu}(X, U^{\nu}): \mathbb{R}^{\bar{n}+\bar{m}^{\nu}}$ $\rightarrow \mathbb{R}$. It depends on player $\nu$'s control inputs $U^{\nu}$ as well as on the state trajectory $X$, which is shared with all the other players. The goal of player $\nu$ is to select a strategy $U^{\nu}$ and a state trajectory $X$ that minimizes the cost function $J^{\nu}$. Naturally, the choice of state trajectory $X$ is constrained by the other players' strategies $U^{-\nu}$ and the dynamics of the system via Equation \ref{eq:dynamics}. In addition, the strategy $U^{\nu}$ must respect a set of constraints that depends on the state trajectory $X$ as well as on the other players strategies $U^{-\nu}$ (e.g., collision-avoidance constraints). We express this with a concatenated set of inequality constraints $C:\mathbb{R}^{\bar{n}+\bar{m}} \rightarrow \mathbb{R}^{n_c}$. \hl{We model the behavior of the players by assuming that each player chooses inputs to solve its own constrained optimization problem, formally,}
    
    \begin{equation}
        \begin{split}
        \min \quad  &J^{1}(X, U^1) \\[-7pt]
        X, U^1 \:\:\: & \\[-1pt]
        \text{s.t.} \quad &D(X,U) = 0, \\
        & C(X,U) \leq 0, \\
        \end{split}
    \quad \quad \ldots \quad \quad
        \begin{split}
        \min \quad  &J^{M}(X, U^M) \\[-7pt]
        X, U^M \:\:\: & \\[-1pt]
        \text{s.t.} \quad &D(X,U) = 0, \\
        & C(X,U) \leq 0, \\
        \end{split}
    \label{pb:gnep}
    \end{equation}
    
    \textbf{\hl{Generalized Nash Equilibrium Problem}} --- \hl{A Nash equilibrium is reached when each of these $M$ coupled optimization problems reaches an optimal point.} 
    This set of $M$ Problems (\ref{pb:gnep}), forms a GNEP because of the constraints that couple the strategies of all the players. A solution of this GNEP (a generalized Nash equilibrium), is a vector $\hat{U}$ such that, for all $\nu = 1, \ldots, M$, $\hat{U}^{\nu}$ is a solution to (\ref{pb:gnep}) with the other players' strategies fixed to $\hat{U}^{-\nu}$. This means that at an equilibrium point $\hat{U}$, no player can decrease their cost by unilaterally changing their strategy $U^{\nu}$ to any other feasible point. 
    
    When solving for a generalized Nash equilibrium of the game, $U$, we identify open-loop Nash equilibrium trajectories, in the sense that the whole trajectory $U^{\nu}$ is the best response to the other players' strategies $U^{-\nu}$ given the initial state of the system $x_0$. Thus, the control signal is a function of time, not of the current state of the system\footnote{One might also explore solving for feedback Nash equilibria, where the strategies are functions of the state of all agents.  This is an interesting direction for future work.} $x_k$. However, one can repeatedly resolve the open-loop game as new information is obtained over time to obtain a policy that is closed-loop in the model-predictive control sense, as demonstrated in Section \ref{sec:mpc}. This formulation is general enough to comprise multi-player dynamic games with nonlinear constraints on the states and control inputs. Practically, in the context of autonomous driving and other scenarios in multi-robot autonomy, the cost function $J^{\nu}$ encodes the objective of player $\nu$, while the concatenated set of constraints, $C$, includes collision constraints coupled between players. We assume differentiability of the constraints and twice differentiability of the cost functions.

\section{Augmented Lagrangian Formulation}
    We propose an algorithm to solve the previously defined GNEP in the context of trajectory optimization. We express the condition that players are acting optimally to minimize their cost functions subject to constraints as an equality constraint. To do so, we first derive the augmented Lagrangian associated with Problem (\ref{pb:gnep}) solved by each player. Then, we use the fact that, at an optimal point, the gradient of the augmented Lagrangian is null \cite{Bertsekas2014}. Therefore, at a generalized Nash equilibrium point, the gradients of the augmented Lagrangians of all players must be null. Concatenating this set of $M$ equality constraints with the dynamics equality constraints, we obtain a set of equations that we solve using a quasi-Newton root-finding algorithm. 
    
    \setcounter{subsection}{0}
    \subsection{Individual Optimality}
    First, without loss of generality, we suppose that the vector $C$ is actually the concatenated set of inequality and equality constraints, i.e., $C = [C_i^T \, C_e^T]^T \in \mathbb{R}^{n_{ci}+n_{ce}}$, where $C_i \leq 0$ is the vector of inequality constraints and $C_e = 0$ is the vector of equality constraints \hl{(e.g. terminal state equality constraints).} 
    To embed the notion that each player is acting optimally, we formulate the augmented Lagrangian associated with Problem (\ref{pb:gnep}) for player $\nu$. The dynamics constraints are handled with the Lagrange multiplier term $\mu^{\nu} \in \mathbb{R}^{\bar{n}}$, while the other constraints are dealt with using both a multiplier and a quadratic penalty term specific to the augmented Lagrangian formulation. As a motivation for this differential treatment; one typically handles inequality and highly nonlinear equality constraints with an augmented Lagrangian formulation for its improved robustness. We denote by $\lambda \in \mathbb{R}^{n_c}$ the Lagrange multipliers associated with the vector of constraints $C$; $\rho \in \mathbb{R}^{n_c}$ is a penalty weight;
    \begin{align}
        L^{\nu}(X, U) = J^{\nu} + {\mu^{\nu}}^T D + {\lambda}^T C + \frac{1}{2} {C}^T I_{\rho}C .
        \label{eq:al}
    \end{align}
    $I_{\rho}$ is a diagonal matrix defined as, 
    \begin{align}
        I_{\rho,kk} &=
        \begin{cases}
            0            & \text{if} \:\:\: C_k(X,U) < 0 \: \land \: \lambda_k = 0, \: k \leq n_{ci}, \\
            \rho_k & \text{otherwise},
        \end{cases}
        \label{eq:penalty_logic}
    \end{align}
    where $k=1, \ldots, n_{ci}+n_{ce}$ indicates the $k^{\mathrm{th}}$ constraint. It is important to notice that the Lagrange multipliers $\mu^{\nu}$ associated with the dynamics constraints are specific to each player $\nu$, but the Lagrange multipliers and penalties $\lambda$ and $\rho$ are common to all players. 
    Given the appropriate Lagrange multipliers $\mu^{\nu}$ and  $\lambda$, the gradient of the augmented Lagrangian with respect to the individual decision variables $\nabla_{X,U^{\nu}} \: L^{\nu} = G^{\nu}$ is null at an optimal point of Problem (\ref{pb:gnep}). The fact that player $\nu$ is acting optimally to minimize $J^{\nu}$ under the constraints $D$ and $C$ can therefore be expressed as follows, 
    \begin{align}
        \nabla_{X, U^{\nu}} \: L^{\nu}(X, U, \mu^{\nu}) = G^{\nu}(X, U, \mu^{\nu}) = 0.
    \end{align}
    It is important to note that this equality constraint preserves coupling between players since the gradient $G^{\nu}$ depends on the other players' strategies $U^{-\nu}$. 
    
    \subsection{Root-Finding Problem}
    At a generalized Nash equilibrium, all players are acting optimally and the dynamics constraints are respected. Therefore, to find an equilibrium point, we have to solve the following root-finding problem,

    \begin{mini}[2]
    {X, U, \mu}{0, \quad \quad \quad \quad \quad \quad \quad \quad \quad \quad \quad \quad \quad \quad}{}{}
    \addConstraint{G^{\nu}(X, U, \mu^{\nu}) = 0, \:\:\: \forall \:\: \nu \in \{1, \ldots, M\}}
    \addConstraint{D(X,U) = 0}
    \label{pb:search}.
    \end{mini}
    \hl{We note that the set of constraints, $C$, is embedded in $G$ and implicitly handled through augmented Lagrangian penalties. Therefore, we do not need to add the primal feasibility constraint, $C(X,U) \leq 0$, to Problem }\ref{pb:search}.
    We use Newton's method to solve the root-finding problem. We denote by $G$ the concatenation of the augmented Lagrangian gradients of all players and the dynamics constraints, $G(X,U,\mu) = [({G^1})^T, \ldots,$ $ ({G^M})^T, D^T]^T $, where $\mu = [(\mu^1)^T, \ldots, (\mu^M)^T]^T \in \mathbb{R}^{\bar{n}M}$. We compute the first-order deri-vative of $G$ with respect to the primal variables $X,U$ and the dual variables $\mu$ that we concatenate in a single vector $y = [(X)^T, (U)^T, (\mu)^{T}]$,
    \begin{align}
        H = \nabla_{X,U,\mu} G = \nabla_y G.
        \label{eq:kkt_jacobian}
    \end{align}
    Newton's method allows us to identify a search direction $\delta y$ in the primal-dual space, 
    \begin{align}
        \delta y = - H^{-1}G.
        \label{eq:newton_step}
    \end{align}
    We couple this search direction with a backtracking line-search \cite{Nocedal2006} given in Algorithm \ref{al:linesearch} to ensure local convergence to a solution using Newton's Method \cite{Nocedal2006} detailed in Algorithm \ref{al:newton}.
    
    \begin{algorithm}
    \caption{Backtracking line-search}\label{al:linesearch}
    \begin{algorithmic}[1]
    \Procedure{LineSearch}{${y}, G, \delta{y}$}
    \State \textbf{Parameters}  
    \State $\alpha = 1$,
    \State $\beta \in (0, 1/2)$,
    \State $\tau \in (0, 1),$
    \State \textbf{Until} {$|| G({y}+\alpha\delta{y}) ||_1 < (1-\alpha\beta) || G({y}) ||_1$} \textbf{do} 
        \State \indent $\alpha \gets \tau \alpha$  
    \State \textbf{return} $\alpha$
    \EndProcedure
    \end{algorithmic}
    \end{algorithm}

    \begin{algorithm}
    \caption{Newton's method for root-finding problem}\label{al:newton}
    \begin{algorithmic}[1]
    \Procedure{Newton'sMethod}{${y}$}
    \State \textbf{Until} {Convergence} \textbf{do}
        \State \indent $G \gets [({\nabla_{X,U^{1}} \: L^{1}})^T, \ldots, ({\nabla_{X, U^{M}} \: L^{M}})^T, D^T]^T$ 
        \State \indent $H \gets \nabla_{y} G$ 
        \State \indent $\delta{y} \gets -H^{-1} G$ 
        \State \indent $\alpha \gets \Call{LineSearch}{{y}, G, \delta{y}}$
        \State \indent ${y} \gets {y} + \alpha \delta{y}$ 
    \State \textbf{return} ${y}$
    \EndProcedure
    \end{algorithmic}
    \end{algorithm}

    \begin{algorithm}
    \caption{ALGAMES solver}\label{al:algames}
    \begin{algorithmic}[1]
    \Procedure{ALGAMES}{${y}_0, \rho_0$}
    \State \textbf{Initialization}  
    \State $\rho \gets \rho^{(0)}, $
    \State $\lambda \gets 0, $   
    \State $\mu^{\nu} \gets 0, $   \hspace*{\fill} $\forall \nu$ 
    \State ${X, U} \gets X^{(0)}, U^{(0)}$ 
    \State \textbf{Until} {Convergence} \textbf{do}
        \State \indent ${y} \gets \Call{Newton'sMethod}{{y}} $ 
        \State \indent $\lambda \gets \Call{DualAscent}{{y}, \lambda, \rho},$ \Comment{Eq. \ref{eq:dual_ascent}}
        \State \indent $\rho \gets \Call{IncreasingSchedule}{\rho},$  \Comment{Eq. \ref{eq:pen_update}}
    \State \textbf{return} ${y}$
    \EndProcedure
    \end{algorithmic}
    \end{algorithm}

    \subsection{Augmented Lagrangian Updates}
    To obtain convergence of the Lagrange multipliers $\lambda$, we update them with a dual-ascent step. This update can be seen as shifting the value of the penalty terms into the Lagrange multiplier terms, 
    \begin{align}
        \lambda_k \leftarrow
        \begin{cases}
            \max(0, \lambda_k + \rho_k C_k(X, U)) 
                & k \leq n_{ci}, \\
            \lambda_k + \rho_k C_k(X,U)   
                & n_{ci} < k \leq n_{ci}+n_{ce}.
        \end{cases}
        \label{eq:dual_ascent}
    \end{align}

    \noindent We also update the penalty weights according to an increasing schedule, with $\gamma > 1$:
    \begin{align}
        \rho_k \leftarrow \gamma \rho_k, \:\:\: \forall k \in \{1, \ldots, n_c\}.
        \label{eq:pen_update}
    \end{align}
    
    \subsection{ALGAMES}
    By combining Newton's method for finding the point where the dynamics is respected and the gradients of the augmented Lagrangians are null with the Lagrange multiplier and penalty updates, we obtain our solver ALGAMES (Augmented Lagrangian GAME-theoretic Solver) presented in Algorithm \ref{al:algames}. The algorithm, which iteratively solves the GNEP, requires as inputs an initial guess for the primal-dual variables $y^{(0)}$ and initial penalty weights $\rho^{(0)}$. The algorithm outputs the open-loop strategies of all players $X,U$ and the Lagrange multipliers associated with the dynamics constraints $\mu$. 
    
    \subsection{Algorithm Complexity} \label{sec:complexity}
    
\begin{figure}[t]
    	\begin{center}
    	\begin{tabular}{c c c}
    		\hline
    	    \specialcellbold{Model} &
    		\specialcellbold{State dim.} & 
    		\specialcellbold{Control dim.} \\
    		\hline
            Quadrotor            &  $12$ & $4$ \\
            Double Integrator 3D &  $6$  & $3$ \\
            Bicycle              &  $4$  & $2$ \\
            Unicycle             &  $4$  & $2$ \\
            Double Integrator 2D &  $4$  & $2$ \\
    		\hline
    	\end{tabular}
    	\end{center}
    
    
        \footnotesize{}
        \vspace{-3mm}
        \hl{\textbf{(a)} ALGAMES is tested on 5 dynamical systems. 
        }
        \vspace{+3mm}
        
        \includegraphics[width=1.00 \linewidth]{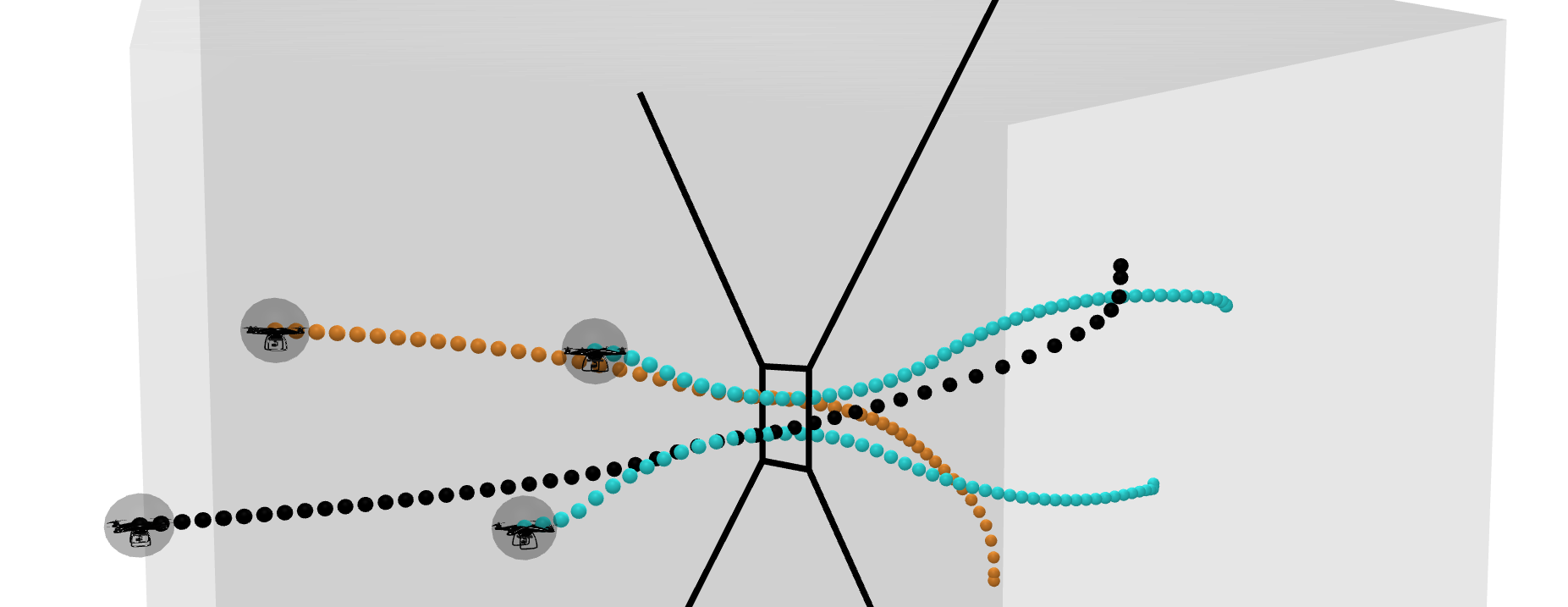}\hfill
        \hl{
        \textbf{(b)} Scenario where drones have to go through a small gap to reach the other side of the room, forcing them to pass through one at a time. 
        }
        \vspace{+3mm}


        \includegraphics[width=1.00 \linewidth, height=0.45\linewidth]{tikz_figure/timings.tikz}\hfill
        \vspace{-1mm}
        \hl{\textbf{(c)} ALGAMES' solve times on the scenario depicted in Figure} \ref{fig:complexity_figure}.b \hl{ averaged across 100 samples. For dynamical systems lying in 2D space (Bicycle, Unicyle and 2D double integrator) the starting points and goal points are projected onto the $XY$ plane.}
        \vspace{+2mm}

        \includegraphics[width=1.00 \linewidth, height=0.6\linewidth]{tikz_figure/convergence.tikz}\hfill
        \hl{\textbf{(d)} Comparison between ALGAMES and IBR 
        on a GNEP with $M$ players, where coupling between players stems from collision avoidance penalties. The convergence rate to a GNE solution is significantly higher for ALGAMES.}
        
        \caption{\hl{We evaluate the scalability of ALGAMES for an increasing number of players and compare its performance to IBR.}}
        \label{fig:complexity_figure}
    \end{figure}

    Following a quasi-Newton approximation of the matrix $H$ \cite{Nocedal2006}, we neglect some of the second-order derivative terms associated with the constraints. \hl{Indeed, these terms involve third-order tensors which require significant amounts of computation per iteration; while only marginally increasing the progress made per iteration.} Therefore, the most expensive part of the algorithm is the Newton step defined by Equation \ref{eq:newton_step}. By exploiting the sparsity pattern of the matrix $H$, we can solve Equation \ref{eq:newton_step} using a back-substitution scheme akin to solving a Riccati recursion with complexity $O(N(n+m)^3)$. The complexity is cubic in the number of states $n$ and the number of control inputs $m$, which are typically linear in the number of players $M$. \hl{Therefore, the complexity of one iteration of the  algorithm is $O(N M^3)$. We evaluate the scalability of ALGAMES for a varying number of players in a drone navigation scenario (Fig.} \ref{fig:complexity_figure}.b \hl{) on 5 different dynamical systems (unicycle, bicycle, quadrotor etc.) with varying number of states and control inputs (Fig. } \ref{fig:complexity_figure}.a \hl{). The timing results are shown in Figure} \ref{fig:complexity_figure}.c.

    \hl{For Newton-type methods, the overall \emph{algorithm complexity} is the same as the \emph{iteration complexity} since the number of iterations required to converge to an optimal solution is independent of the problem size; provided that the initial guess is close to an optimal solution} \cite{Facchinei2007}. \hl{The same reasoning about complexity holds for iLQGames.}    \hl{
    On the other hand, IBR has a smaller iteration complexity, $O(N M)$, it is linear in terms of the number of players. However, the overall algorithm complexity depends on rate of convergence to a Nash equilibrium. In theory, convergence is guaranteed under extremely restrictive conditions} \cite{Facchinei2007}. \hl{In practice, we have observed low convergence rates on a simple problem where the only constraints were linear kinodynamics constraints (Fig. }\ref{fig:complexity_figure}.d). \hl{ ALGAMES is 4 to 10 times faster than IBR for a typical convergence threshold (residual $< 10^{-4}$.) Thus, even with a large number of players, ALGAMES remains significantly faster than IBR.}

    \subsection{Algorithm Discussion}
    Here we discuss the inherent difficulty in solving for Nash equilibria in large problems, and explain some of the limitations of our approach. First of all, finding a Nash equilibrium is a non-convex problem in general. Indeed, it is known that even for single-shot discrete games, solving for exact Nash equilibria is computationally intractable for a large number of players    \cite{DaskalakisEtAlSIAMJournalonComputing08ComplexityOfNash}.  It is, therefore, not surprising that, in our more difficult setting of a dynamic game in continuous space, no guarantees can be provided about finding an exact Nash equilibrium.  Furthermore, in complex interaction spaces, constraints can be highly nonlinear and nonconvex. This is the case in the autonomous driving context with collision-avoidance constraints. In this setting, one cannot expect to find an algorithm working in polynomial time with guaranteed convergence to a Nash equilibrium respecting constraints. On the other hand, \emph{local} convergence of Newton's method to open-loop Nash equilibria has been established in the unconstrained case (that is, starting sufficiently close to the equilibrium, the algorithm will converge to it) \cite{Di2019}. Our approach solves a sequence of unconstrained problems via the augmented Lagrangian formulation. Each of these problems, therefore, has guaranteed \emph{local} convergence. However, as expected, the overall method has no guarantee of global convergence to a generalized Nash equilibrium.  
    
    Second, our algorithm requires an initial guess for the state and control input trajectories $X$, $U$ and the dynamics multipliers $\mu$. Empirically, we observe that choosing $\mu = 0$ and simply rolling out the dynamics starting from the initial state $x_0$ without any control was a sufficiently good initial guess to get convergence to a local optimum that respects both the constraints and the first-order optimality conditions. For a detailed empirical study of the convergence of ALGAMES and its failure cases, we refer to Sections \ref{sec:monte_carlo} and \ref{sec:failure}. 

    Finally, even for simple linear-quadratic games, the Nash equilibrium solution is not necessarily unique. In general, an entire subspace of equilibria exists. In this case, the matrix $H$ in Equation \ref{eq:newton_step} will be singular. In practice, we regularize this matrix so that large steps $\delta y$ are penalized, resulting in an invertible matrix $H$.

\section{Simulations: Design and Setup}
    We choose to apply our algorithm in the autonomous driving context. Indeed, many maneuvers like lane changing, ramp merging, overtaking, and intersection crossing involve a high level of interaction between vehicles. We assume a single car is computing the trajectories for all cars in its neighborhood, so as to find its own trajectory to act safely among the group. We assume that this car has access to a relatively good estimate of the surrounding cars' objective functions. Such an estimate could, in principle, be obtained by applying inverse optimal control on observed trajectories of the surrounding cars. 
    
    In a real application, the car would compute its strategy as frequently as possible in a receding-horizon loop to adapt to unforeseen changes in the environment. We demonstrate the feasibility of this approach on complex driving scenarios where a classical predict-then-plan architecture fails to overcome the ``frozen robot" problem.

    \setcounter{subsection}{0}
    \subsection{Autonomous Driving Problem}
    \paragraph{Constraints}
    Each vehicle in the scene is an agent of the game. Our objective is to find a generalized Nash equilibrium trajectory for all of the vehicles. These trajectories have to be dynamically feasible. The dynamics constraints at time step $k$ are expressed as follows, 
    \begin{align}
        x_{k+1} = f(x_k, u^{1}_k, \ldots, u^{M}_k).
    \end{align}
    We consider a nonlinear unicycle model for the dynamics of each vehicle. A vehicle state, $x^{\nu}_k$, is composed of a 2D position, a heading angle and a scalar velocity. The control input $u^{\nu}_k$ is composed of an angular velocity and a scalar acceleration. 
    In addition, it is critical that the trajectories respect collision-avoidance constraints. We model the collision zone of the vehicles as circles of radius $r$. The collision constraints between vehicles are then simply expressed in terms of the position $\tilde{x}^{\nu}_{k}$ of each vehicle, 
    \begin{align}
        (2r)^2 - || \tilde{x}^{\nu}_{k} - \tilde{x}^{\omega}_{k} ||_2^2 \leq 0, \quad \forall \: \: \nu, \omega \in \{1, \ldots, M\},  \nu \neq \omega.
    \end{align} 
    We also model boundaries of the road to force the vehicles to remain on the roadway. This means that the distance between the vehicle and the closest point, $q$, on each boundary, $b$, has to remain larger than the collision-circle radius, $r$,
    \begin{align}
        r^2 - || \tilde{x}^{\nu}_{k} - q_b||_2^2 \leq 0, \quad \forall \:\: b, \: \forall \: \: \nu \in \{1, \ldots, M\}.
    \end{align}
    
    In summary, based on reasonable simplifying assumptions, we have expressed the driving problem in terms of non-convex and non-linear coupled constraints.  
    
    \paragraph{Cost Function}
    We use a quadratic cost function penalizing the use of control inputs and the distance between the current state and the desired final state $x_f$ of the trajectory. We also add a quadratic penalty on being close to other cars,
    \begin{align}
        J^{\nu}(&X, U^{\nu}) = \sum_{k=1}^{N-1} \frac{1}{2}(x_k - x_f)^T Q (x_k - x_f) + \frac{1}{2}{u^{\nu}_k}^T R u^{\nu}_k \nonumber \\
        &+ \frac{1}{2}(x_N - x_f)^T Q_f (x_N - x_f)  \\
        &+ \sum_{k=1}^{N} \sum_{\omega \neq \nu} 
            \gamma \bigg( \max{\big(0, || \tilde{x}_k^{\nu} - \tilde{x}_k^{\omega} ||_2 - \eta \big)} \bigg)^2, \nonumber
    \end{align}
    $\eta$ controls the distance at which this penalty is ``activated'', and $\gamma$ controls its magnitude. 

    \subsection{Comparison to iLQGames}
    In order to evaluate the merits of ALGAMES, we compare it to iLQGames \cite{Fridovich-Keil2020a} which is a DDP-based algorithm for solving general dynamic games. Both algorithms solve the problem by iteratively solving linear-quadratic approximations that have an analytical solution \cite{Basar1999}. For iLQGames, the augmented objective function $\hat{J}^{\nu}$ differs from the objective function, $J^{\nu}$, by a quadratic term penalizing constraint violations,
    \begin{align}
        \hat{J}^{\nu}(X,U) = J^{\nu}(X,U) + \frac{1}{2} C(X,U)^T I_{\rho} C(X,U).
    \end{align}
    Where $I_\rho$ is defined by, 
    \begin{align}
        I_{\rho,kk} &=
        \begin{cases}
            0            & \text{if} \:\:\: C_k(X,U) < 0 , \: k \leq n_{ci}, \\
            \rho_k & \text{otherwise}.
        \end{cases}
    \end{align}
    Here $\rho$ is an optimization hyperparameter that we can tune to satisfy constraints. For ALGAMES, the augmented objective function, $L^{\nu}$, is actually an augmented Lagrangian, see Equation \ref{eq:al}. The hyperparameters for ALGAMES are the initial value of $\rho^{(0)}$ and its increase rate $\gamma$ defined in Equation \ref{eq:pen_update}.

    \begin{figure}[t]
    \includegraphics[width=8.3cm]{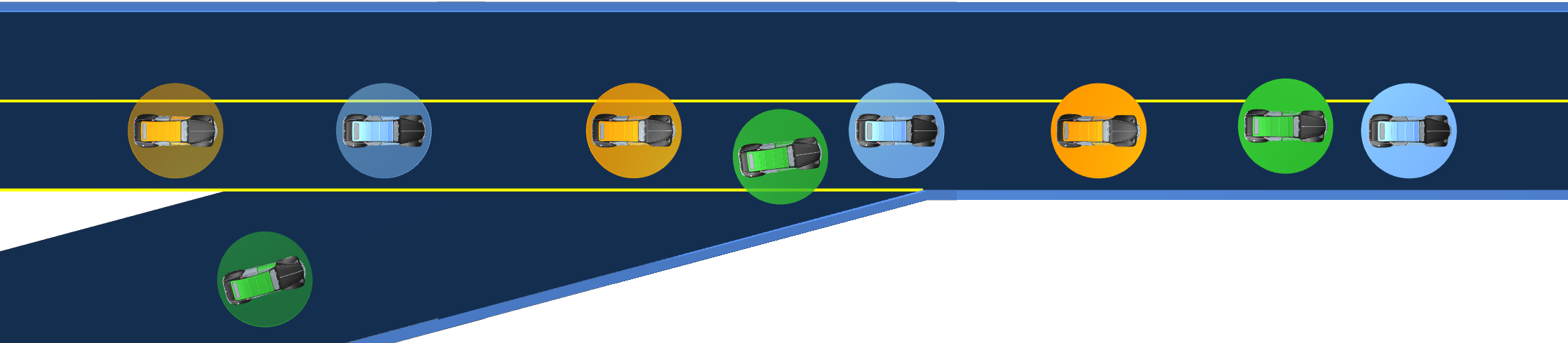}\hfill
    \footnotesize{} \textbf{(a)} On the left, the three cars at their nominal initial state. On the right, the three cars are standing at the desired final state. The green car has successfully merged in between the two other cars. The roadway boundaries are depicted in light blue.
    \vspace{+2mm}
    
    \includegraphics[width=8.3cm]{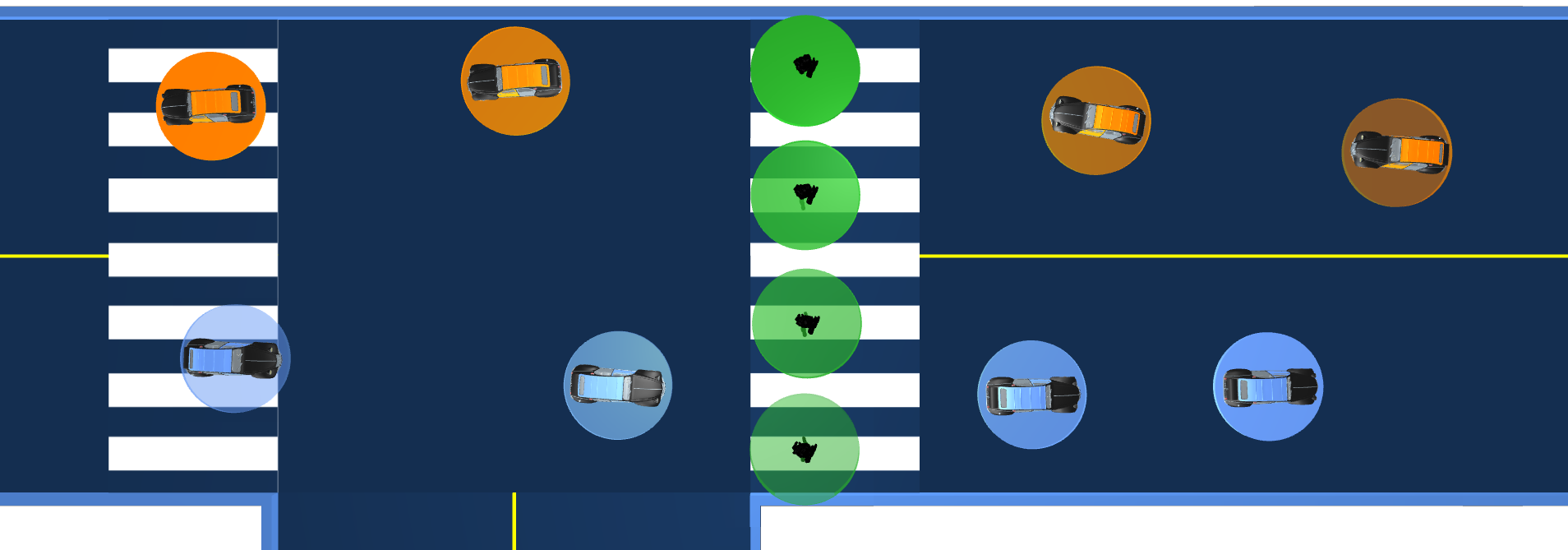}\hfill
    \textbf{(b)} The blue car starts on the left and finishes on the right. The orange does the opposite. The pedestrian with the green collision-avoidance circle crosses the road from the bottom to the top of the image.
    
    \caption{Two driving environments are considered: a ramp merging scenario (top) and an intersection crossing scenario (bottom). 
    }
    \label{fig:ramp_merging & intersection}
    \end{figure}

    
    \begin{figure}[t]
        \begin{center}
        \begin{tabular}{|c|c|c|c|}
        \hline
        Scenario & \# Players & ALGAMES & iLQGames\\
        \hline
        Ramp  & 2 & $\mathbf{38 \pm 10 ms}$ & $104 \pm 23 ms$\\
        Merging & 3 & $\mathbf{89 \pm 14 ms}$ & $197 \pm 15 ms$\\
          & 4 & $860 \pm 251 ms$ & $\mathbf{705 \pm 209 ms}$\\
        \hline
          & 2 & $\mathbf{50 \pm 11 ms}$ & $752 \pm 168 ms$\\
        Intersection & 3 & $\mathbf{116 \pm 22 ms}$ & $362 \pm 93 ms$\\
          & 4 & $\mathbf{509 \pm 33 ms}$ & $1905 \pm 498 ms$\\
        \hline 
        \end{tabular}\\
        \end{center}
        \footnotesize{} \textbf{(a)} For each scenario and each number of players, we run each solver 100 times to estimate the mean solve time and its standard deviation.
        \vspace{+4mm}
        
        \includegraphics[width=.49\linewidth, height=.32\linewidth]{tikz_hist/const_dir.tikz}\hfill
        \includegraphics[width=.49\linewidth, height=.32\linewidth]{tikz_hist/const_ilq.tikz}
        \includegraphics[width=.49\linewidth, height=.32\linewidth]{tikz_hist/time_dir.tikz}\hfill
        \includegraphics[width=.49\linewidth, height=.32\linewidth]{tikz_hist/time_ilq.tikz}
        \includegraphics[width=.49\linewidth, height=.32\linewidth]{tikz_hist/iter_dir.tikz}\hfill
        \includegraphics[width=.49\linewidth, height=.32\linewidth]{tikz_hist/iter_ilq.tikz}
        \textbf{(b)} Monte Carlo analysis with 1000 randomly sampled initial states of ALGAMES on the left and iLQGames on the right. The top plots indicate the largest constraint violation of the solution at the end of the solve, $\epsilon \geq 0$ (smaller $\epsilon$ means constraints are better satisfied). Middle plots show the solve time. The bottom left and right plots displays the number of Newton steps and the number of Riccati backward passes executed during the solve of ALGAMES and iLQGames respectively.
        \vspace{+2mm}
        \begin{center}
        \begin{tabular}{|c c c c|} 
            \hline
            Scenario     & Frequency & $\mathbb{E}[\delta t]$ & $\mathbb{\sigma}[\delta t]$ \\ [0.5ex] 
            \hline
            Ramp Merging & 69 Hz & 14 ms & 72 ms \\ 
            Intersection & 66 Hz & 15 ms & 66 ms \\
            \hline
        \end{tabular}\\
        \end{center}
        \vspace{+2mm}
        \textbf{(c)} Running the Model-Predictive Control (MPC) implementation of ALGAMES 100 times on both scenarios, we obtain the mean update frequency of the MPC as well as the mean and standard deviation of $\delta t$, the time required to update the MPC plan.
        
        \caption{We compare ALGAMES and iLQGames, both in terms of solve time (top) and ability to reliably enforce constraints (middle). Additionally, we evaluate the update frequency of ALGAMES in the MPC setting (bottom).}
        \label{fig:benchmark & monte_carlo & mpc_table}
    \end{figure}

    \subsection{Timing Experiments}
    We evaluate the performance of both algorithms in two scenarios (see Figure \ref{fig:ramp_merging & intersection}) with the number of players varying from two to four. To compare the speed of both algorithms, we set the termination criterion as a threshold on constraint violations $C \leq 10^{-3}$. The timing results averaged over 100 samples are presented in Table \ref{fig:benchmark & monte_carlo & mpc_table}.a. First, we notice that both algorithms achieve real-time or near-real-time performance on complex autonomous driving scenarios (the horizon of the solvers is fixed to $5s$).
    
    We observe that the speed performance of ALGAMES and iLQGames are comparable in the ramp merging scenario. For this scenario, we tuned the value of the penalty for iLQGames to $\rho = 10^2$. Notice that for all scenarios the dimensions of the problem are scaled so that the velocities and displacements are all the same order of magnitude. For the intersection scenario, we observe that the two-player and four-player cases both have much higher solve times for iLQGames compared to the 3-player case. Indeed, in those two cases, we had to increase the penalty to $\rho = 10^3$, otherwise the iLQGames would plateau and never reach the constraint satisfaction criterion. This, in turn, slowed the algorithm down by decreasing the constraint violation convergence rate. 
    
    \subsection{Discussion}
    The main takeaway from these experiments is that, for a given scenario, it is generally possible to find a suitable value for $\rho$ that will ensure the convergence of iLQGames to constraint satisfaction. With higher values for $\rho$, we can reach better constraint satisfaction at the expense of slower convergence rate. In the context of a receding horizon implementation (MPC), finding a good choice of $\rho$ that would suit the whole sequence of scenarios encountered by a vehicle could be difficult. In contrast, the same hyperparameters $\rho^{(0)}=1$ and $\gamma =10$ were used in ALGAMES for all the experiments across this paper. This supports the idea that, thanks to its adaptive penalty scheme, ALGAMES requires little tuning. 
    
    While performing the timing experiments, we also noticed several instances of oscillatory behavior for iLQGames. The solution would oscillate, preventing it from converging. This happened even after an adaptive regularization scheme was implemented to regularize iLQGames' Riccati backward passes. Oscillatory behavior was not seen with ALGAMES. We hypothesize that this is due to the dual ascent update coupled with the penalty logic detailed in Equations \ref{eq:dual_ascent} and \ref{eq:penalty_logic}, which add hysteresis to the solver.


    \subsection{Monte Carlo Analysis} \label{sec:monte_carlo}
    To evaluate the robustness of ALGAMES, we performed a Monte Carlo analysis of its performance on a ramp merging problem. First, we set up a roadway with hard boundaries as pictured in Fig. \ref{fig:ramp_merging & intersection}.a. We position two vehicles on the roadway and one on the ramp in a collision-free initial configuration. We choose a desired final state where the incoming vehicle has merged into the traffic. Our objective is to generate generalized Nash equilibrium trajectories for the three vehicles. These trajectories are collision-free and cannot be improved unilaterally by any player. To introduce randomness in the solving process, we apply a random perturbation to the initial state of the problem. Specifically, we perturb $x_0$ by adding a uniformly sampled noise. This would typically correspond to displacing the initial position of the vehicles by $\pm 1 m$, changing their initial velocity by $\pm 3\%$ and their heading by $\pm 2.5^\circ$. 
    
    We observe in Figure \ref{fig:benchmark & monte_carlo & mpc_table}.b, that ALGAMES consistently finds a satisfactory solution to the problem using the same hyperparameters $\rho^{(0)} =1$ and $\gamma = 10$. Out of the 1000 samples $99.5\%$ converged to constraint satisfaction $C \leq 10^{-3}$ while respecting the optimality criterion $||G||_1 < 10^{-2}$. By definition, $||G||_1$ is a merit function for satisfying optimality and dynamics constraints.  We also observe that the solver converges to a solution in less than $0.2 s$ for $96\%$ of the samples. 
    These empirical data tend to support the fact that ALGAMES is able to solve the class of ramp merging problem quickly and reliably. 
    
    For comparison, we present in Figure \ref{fig:benchmark & monte_carlo & mpc_table}.b the results obtained with iLQGames. We apply the same constraint satisfaction criterion $C \leq 10^{-3}$. We fixed the value of the penalty hyperparameter $\rho$ for all the samples as it would not be a fair comparison to tune it for each sample. Only 3 samples did not converge with iLQGames, this is a performance comparable to ALGAMES for which 5 samples failed to converge. However, we observe that iLQGames is 3 times slower than ALGAMES with an average solve time of $350$ ms compared to $110$ ms and require on average 4 times more iterations (9 against 41).

    \subsection{Solver Failure Cases}  \label{sec:failure}
    The Monte Carlo analysis allows us to identify the typical failure cases of our solver, i.e. the cases where the solver does not satisfy the constraints or the optimality criterion.
    Typically in such cases, the initial guess, which consists of rolling out the dynamics with no control, is far from a reasonable solution. Since the constraints are ignored during this initial rollout, the car at the back can overtake the car at the front by driving through it. This creates an initial guess where constraints are strongly violated. Moreover, we hypothesize that the tight roadway-boundary constraints tend to strongly penalize solutions that would 'disentangle' the car trajectories because they would require large boundary violation at first. Therefore, the solver gets stuck in this local optimum where cars overlap each other. Sampling several initial guesses with random initial control inputs and solving in parallel could reduce the occurrence of these failure cases. Also, being able to detect, reject, and re-sample initial guesses when the initial car trajectories are strongly entangled could also improve the robustness of the solver.

\section{Non-Uniqueness of Nash Equilibria}
    \label{sec:nu}
    A Nash equilibrium corresponds to a situation where all players are acting optimally given the other players' strategies. This is a way for players to compete in a coordinated fashion without communication. However, if the Nash equilibrium is non-unique, the coordination is ambiguous and players have to decide individually which Nash equilibrium to follow. This can lead to inconsistencies. The non-unique-ness of Nash equilibrium solutions has been observed in practical robotics applications such as autonomous driving \cite{Peters2020}. Peters et al. identified isolated clusters of solutions in unconstrained Nash equilibrium problems and proposed an estimation method to improve players' coordination. In this section, we detail several underlying causes of non-uniqueness that arise in practical robotics scenarios. Additionally, we present the behavior of ALGAMES in such circumstances. 
    
    \subsection{Linear-Quadratic Dynamic Games}
        Linear-quadratic (LQ) dynamic games are an important building block for optimization algorithms relying on sequential-quadratic approximations such as ALGAMES or iLQGames \cite{Fridovich-Keil2020a}. The conditions for the existence and uniqueness of a Nash equilibrium have been extensively studied \cite{Basar1976}, \cite{Abraham2019}. In the continuous-time setting, Eisele characterized the different solution regimes for the LQ game, including non-existence and non-uniqueness \cite{Eisele1982}. In the discrete-time setting, the open-loop Nash equilibrium problem is equivalent to a static quadratic game (i.e., a one-step game). For such problems, the Nash equilibrium solutions can either be non-existant, can form an affine subspace, or be a single point in the case of a unique solution.
        \paragraph{Proof Sketch}: We focus on the two-player case, the result can easily be extended to the $M$-player case. We denote, $s^{\nu}$ and $J^{\nu}$, the strategy and quadratic cost function of player $\nu$,
        \begin{align}
            J^{\nu}(s^1, s^2) = \frac{1}{2}
                \begin{bmatrix}
                    s^1 \\
                    s^2
                \end{bmatrix}^T
                \begin{bmatrix}
                    Q^{\nu}_{1,1} & Q^{\nu}_{1,2} \\
                    {Q^{\nu}_{2,1}} & Q^{\nu}_{2,2}
                \end{bmatrix}
                \begin{bmatrix}
                    s^1 \\
                    s^2
                \end{bmatrix}
                +
                \begin{bmatrix}
                    s^1 \\
                    s^2
                \end{bmatrix}^T
                \begin{bmatrix}
                    q^{\nu}_{1} \\
                    q^{\nu}_{2}
                \end{bmatrix}
                + 
                c^{\nu}.
        \end{align}
        The first-order necessary conditions for optimality of a Nash equilibrium, $\hat{e} = ({\hat{s}^1}, {\hat{s}^2})$, can be written as an affine equation, 
        \begin{align}
        \label{eq:affine_subspace}
            \begin{bmatrix}
                Q^{1}_{1,1} + Q^2_{2,1} & Q^{2}_{2,2} + Q^1_{1,2}
            \end{bmatrix}
            \begin{bmatrix}
                \hat{s}^1 \\
                \hat{s}^2
            \end{bmatrix}
            + 
            \begin{bmatrix}
            q^1_1 \\
            q^2_2
            \end{bmatrix} = 0.
        \end{align}
        The second-order necessary conditions are independent of the Nash equilibrium point considered. They require positive semi-definiteness of the matrices $Q^{\nu}_{\nu,\nu}$, for all $\nu \in \{1, 2\}$. Therefore, any point, $e$, in the affine subspace defined by Equation \ref{eq:affine_subspace}, will respect both the first-order and second-order necessary conditions for optimality. $\blacksquare$ 
        
        In case of a unique Nash Equilibrium, ALGAMES converges in one Newton iteration to the solution. When the Nash equilibrium solutions form an affine subspace, ALGAMES converges to the point in the subspace closest to the initial guess. This is due to the regularization added to the Jacobian of the KKT condition, $H$, defined in Equation \ref{eq:kkt_jacobian}. 
    
        \subsection{Isolated Nash Equilibria}
\begin{figure}[t]
            \includegraphics[width=8.3cm]{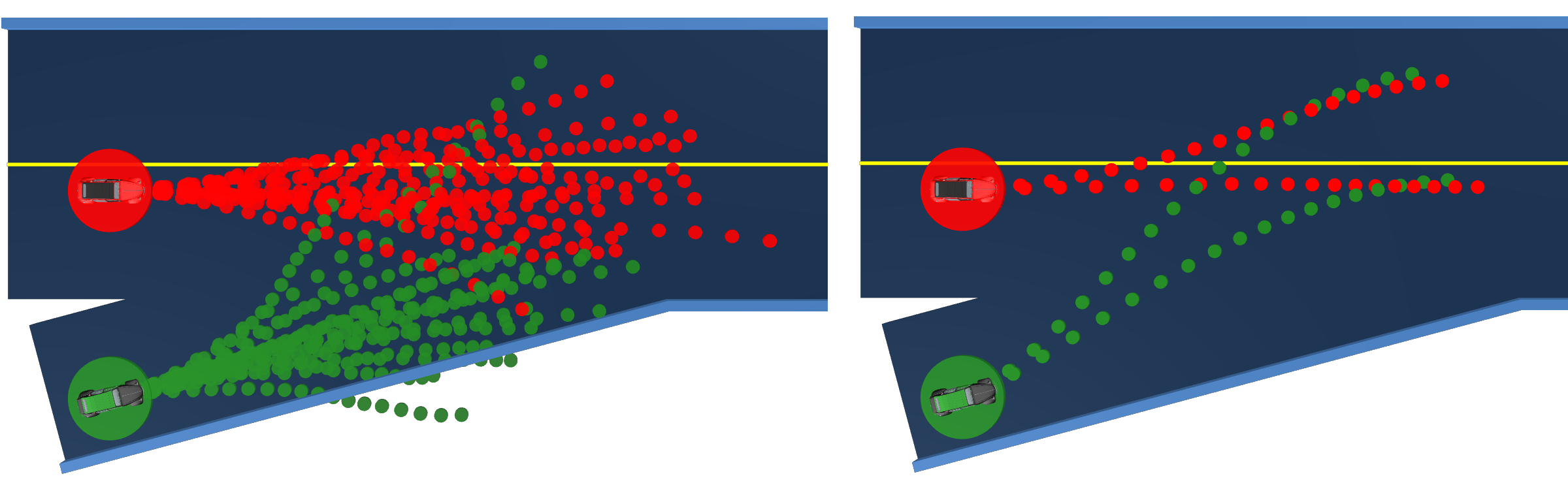}
            \footnotesize{} \textbf{(a)} We initialize ALGAMES with 20 randomly-sampled and dynamically-feasible trajectories (left). For this unconstrained but non-convex problem, the solver converges to two ``topologically'' different solutions (right). Either the green car goes first and merges into the bottom lane, or the green car lets the orange car go first and merges into the top lane. 
            \vspace{+2mm}
            
            \includegraphics[width=8.5cm]{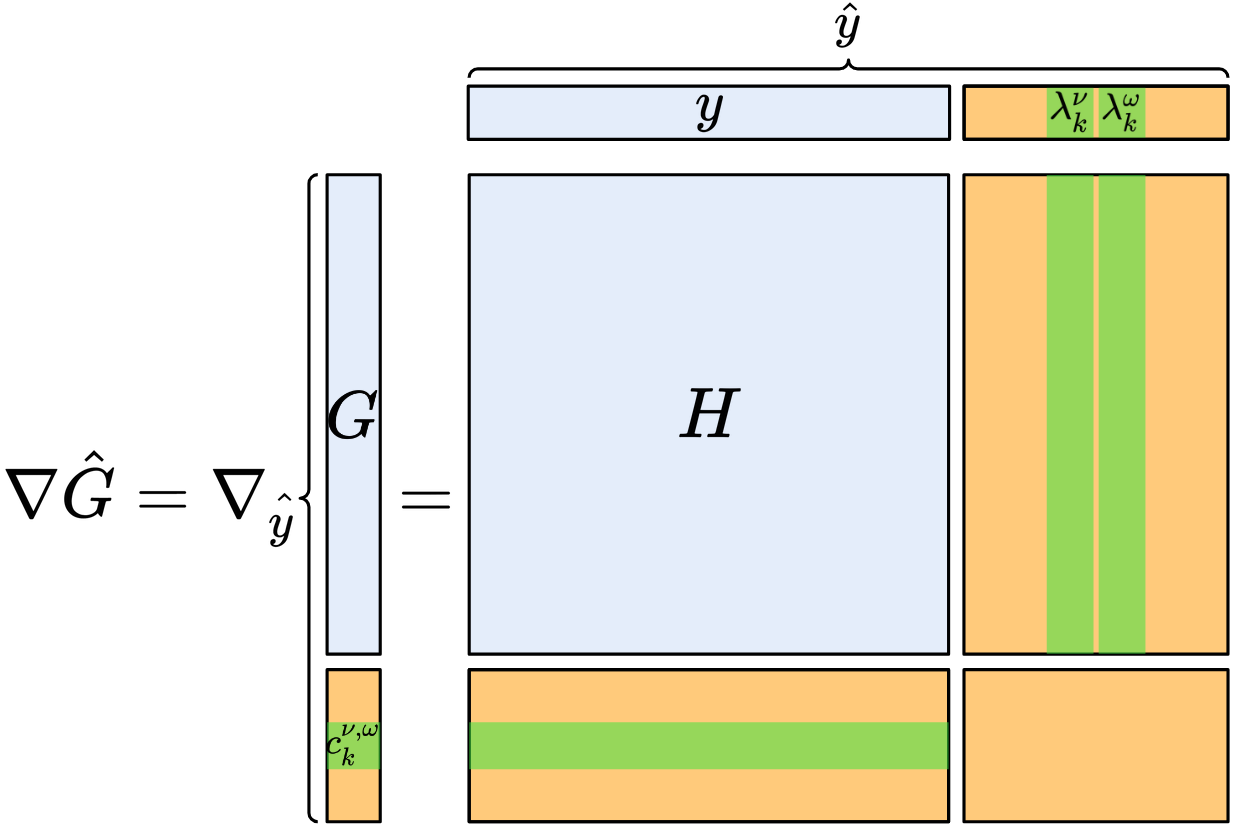}\hfill
            \textbf{(b)} The main block, $\nabla_y G = H$, depicted in light blue is a square matrix. Each active collision-avoidance constraint, $c_k^{\nu,\omega}$, adds a row to the Jacobian matrix, $\hat{H} = \nabla_{\hat{y}} \hat{G}$. Simultaneously, it introduces two Lagrange multipliers, $\lambda_k^{\nu}$ and $\lambda_k^{\omega}$, adding two columns to the Jacobian, $\hat{H}$. Consequently, the identification of the search direction in Newton's method, defined by Equation \ref{eq:newton_step}, is turned into an underdetermined linear system of equations, $\delta \hat{y} = -\hat{H}^{-1} \hat{G}$.

            \caption{We illustrate isolated Nash equilibria (top), and non-isolated Nash equilibrium solutions stemming from an underdetermined KKT system (bottom).}
            \label{fig:unconstrained_local_min & fat_matrix}
        \end{figure}

        We have seen that an LQ game can generate an affine Nash equilibrium subspace. Thus, it cannot lead to multiple isolated Nash equilibria. However, in general, a dynamic game can admit multiple isolated Nash equilibria as highlighted by Peters et al. \cite{Peters2020}. 
        In unconstrained autonomous driving scenarios, they generally appear when collision-avoidance costs are introduced. These costs are non-convex and introduce a coupling between the players' strategies. Typically, these isolated Nash equilibria correspond to ``topologically'' different driving strategies. For instance, in a ramp merging scenario, the merging vehicle can merge in front of or behind an incoming vehicle (Figure \ref{fig:unconstrained_local_min & fat_matrix}.a). The equilibrium point to which ALGAMES converges is typically the closest to the initial guess, thanks to the regularization scheme. This is a desirable property, especially in the MPC setting , because it prevents the re-planned trajectory from oscillating between different Nash equilibria. 


    \subsection{Generalized Nash Equilibrium}
    \label{sec:fat_matrix}
    Identifying the solution set of a GNEP remains a major challenge as pointed out by Fisher et al. in an extensive survey \cite{Fischer2014}. In general, the solution set of the GNEP can be constituted of one or many isolated points or even non-isolated points. 
    Theoretical results in this domain often rely on strong assumptions, such as convexity of the feasible set, absence of shared constraints, or decoupled cost functions \cite{Dreves2018}. All these assumptions could be violated in a typical robotic scenario. Indeed, collision-avoidance constraints are shared and non-convex. Similarly, collision-avoidance costs or congestion terms introduce coupling between the players' costs. 
    
    
    We explore the structure of the generalized Nash equilibrium (GNE) solutions in the presence of shared collision-avoidance constraints. We denote, $c^{\nu,\omega}_k : \mathbb{R}^n \rightarrow \mathbb{R}$, the collision-avoidance constraint between player $\nu$ and player $\omega$ at time step $k$. For each collision-avoidance constraint $c^{\nu,\omega}_k$, we introduce two Lagrange multipliers $\lambda_k^{\nu} \in \mathbb{R}$ and $\lambda_k^{\omega} \in \mathbb{R}$; one for each player. We remark that, for a single constraint, we add two Lagrange multipliers. 
    We denote, $N_c$, the number of collision-avoidance constraints. By concatenating these constraints with the residual vector $G$, we add $N_c$ entries and $N_c$ rows to its Jacobian. We denote $\hat{G}$ and $\hat{H}$ the ``augmented'' residual vector and Jacobian matrix. We need to differentiate the residual $\hat{G}$, with respect to the $2 N_c$ Lagrange multipliers associated with the collision constraints. This adds $2 N_c$ columns to the Jacobian $\hat{H}$. Thus, the Jacobian is an underdetermined linear system, with $N_c$ more columns than rows (Figure \ref{fig:unconstrained_local_min & fat_matrix}.b). However, only active collision-avoidance constraints should be included in the Jacobian.
    Therefore, the Jacobian only has $N_a$ more columns than rows, where $N_a$ denotes the number of active collision-avoidance constraints. Thus, the nullspace of the underdetermined linear system $\hat{H}$ is at least of dimension $N_a$ (Figure \ref{fig:unconstrained_local_min & fat_matrix}.b).
    Consequently, the solution set of a GNEP can potentially be composed of non-isolated points and could span in multiple dimensions locally around a known equilibrium point.
    
\begin{figure}
        \includegraphics[width=8.3cm]{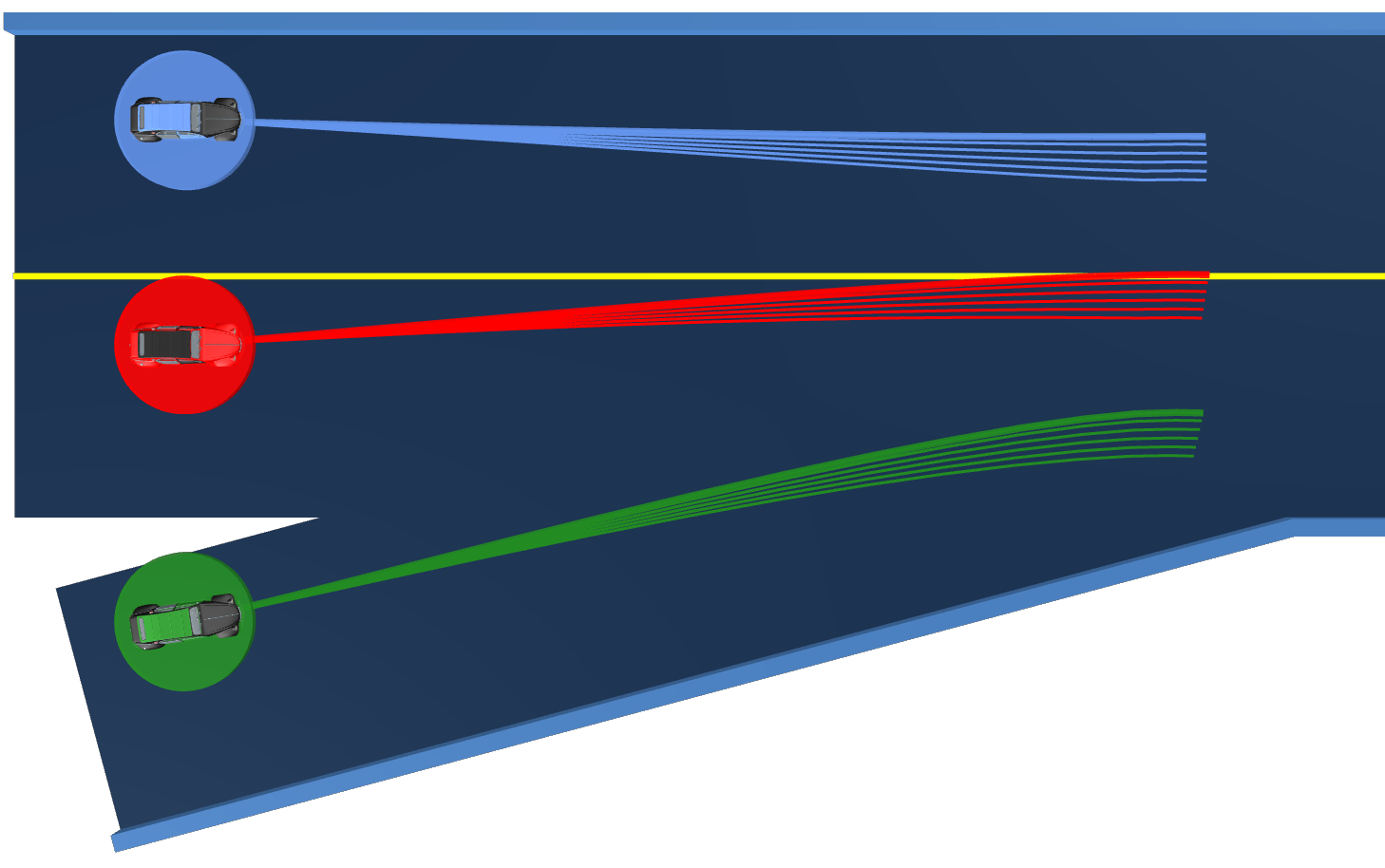}
        \footnotesize{} \textbf{(a)} At a Nash equilibrium point, where collision constraints are active; we compute the nullspace of the KKT conditions' Jacobian depicted in Figure \ref{fig:unconstrained_local_min & fat_matrix}.b. We choose one vector in this nullspace and explore the Nash equilibrium subspace in this direction. We take a small step in the chosen direction, then project back onto the generalized Nash Equilibrium subspace by resolving the GNEP using ALGAMES.
        \vspace{+2mm}
        
        \includegraphics[width=8.3cm]{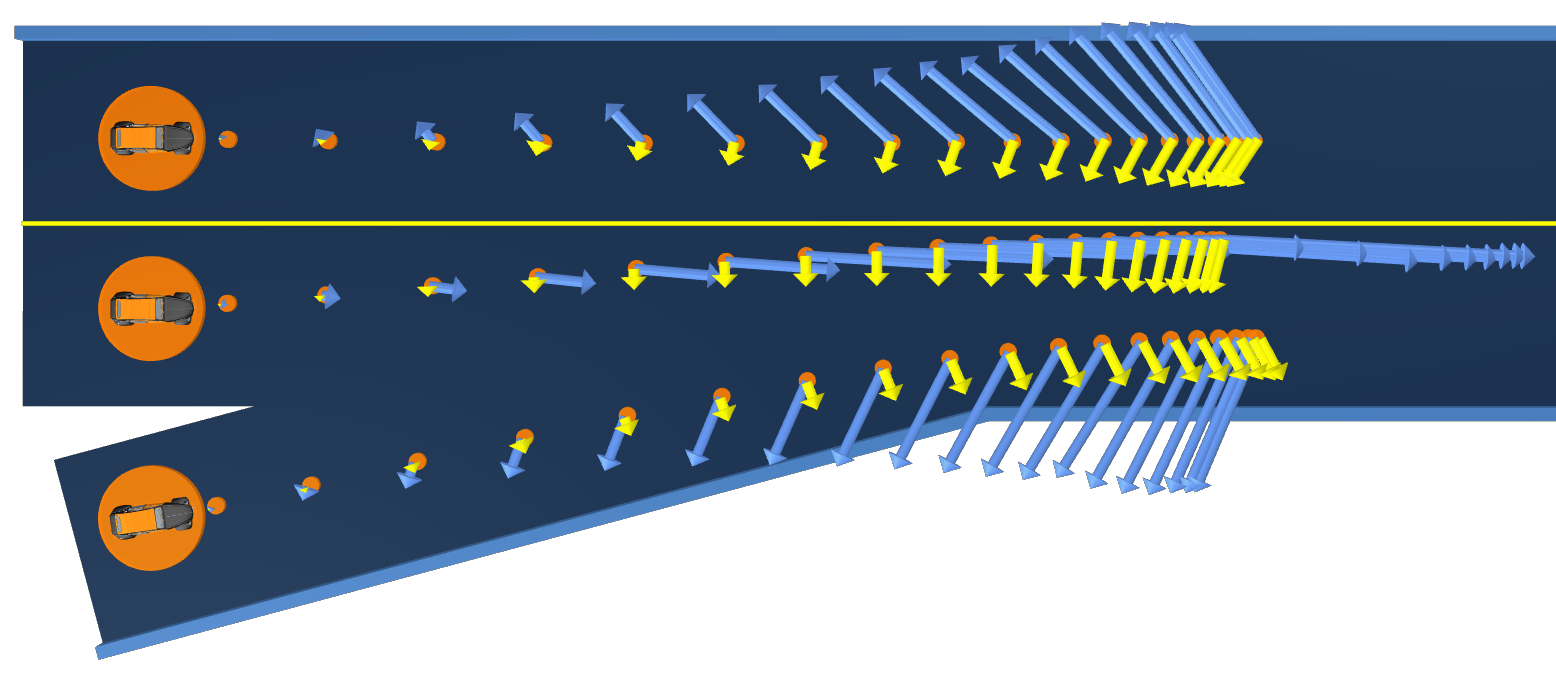}
        \textbf{(b)} At a Nash equilibrium point, we compute the nullspace of the KKT system depicted in Figure \ref{fig:unconstrained_local_min & fat_matrix}.b. Given a basis for the nullspace, for each basis vector, we disturb the nominal solution in its direction. Then, each disturbed solution is projected back onto the generalized Nash equilibrium subspace. Finally, we perform a principal component analysis (PCA) to visualize the vectors responsible for the largest part of the variance in the trajectory of the vehicles. The yellow and blue arrows represent the eigenvectors corresponding to the two largest eigenvalues obtained with the PCA (Figure \ref{fig:pca & active_constraints}.c). The objective for all cars is to drive in the yellow centerline. Thus, moving along the direction of the blue vector would favor the middle car and penalize the top and bottom cars. Following the yellow vector would advantage the top vehicle to the detriment of the middle and bottom player. 
        \vspace{+2mm}
        
        \includegraphics[width=8.3cm, height=3.5cm]{tikz_figure/pca.tikz}
        \textbf{(c)} Eigenvalues obtained with the PCA performed on the disturbed solution vectors.

        \caption{We explore the generalized equilibrium subspace in one direction and obtain a sequence of GNE (top). Additionally, we represent the two principal vectors along which the trajectories can evolve while remaining a GNE (middle \& bottom).}
        \label{fig:pca & active_constraints}
    \end{figure}

    We explore this nullspace at a Nash equilibrium point by slightly disturbing the current generalized Nash equilibrium in one of the nullspace's directions (Figure \ref{fig:pca & active_constraints}.a). We obtain a continuum of GNE. Additionally, Figures \ref{fig:pca & active_constraints}.b and \ref{fig:pca & active_constraints}.c, present the two main directions in which the solution can drift while remaining a GNE. The nullspace was of dimension $17$, which corresponds to the number of active constraints at the equilibrium point. Yet, we notice that most of the trajectory variability is captured by a limited number of eigenvectors. We remark that the two principal eigenvectors have an elegant interpretation: they both favor one vehicle over the others. Additionally, they nicely show how disturbing the trajectory of one player influences the trajectories of the other players through the collision constraints. Finally, by combining these two eigenvectors and stepping in the resulting direction, one could favor any of the three vehicles.
    

    \subsection{ALGAMES' Convergence to Normalized Nash Equilibrium}
    We observe that the multipliers $\lambda_k^{\nu}$ and $\lambda_k^{\omega}$ associated with the shared constraint, $c_k^{\nu,\omega}$, are equal at every iteration of the solver. Indeed, we can assume that the multipliers are initialized with the same value (typically zero). Moveover, these multipliers are updated with identical dual ascent updates, defined in Equation \ref{eq:dual_ascent}, 
    \begin{align}
        & \lambda_k^{\nu (0)} = \lambda_k^{\omega (0)} \\
        & \lambda_k^{\nu (t+1)} = \max\big(0, \lambda_k^{\nu (t)} + \rho_k c_k^{\nu,\omega}(x_k)\big) \\
        & \lambda_k^{\omega (t+1)} = \max\big(0, \lambda_k^{\omega (t)} + \rho_k c_k^{\nu,\omega}(x_k)\big) \\
        \implies & \lambda_k^{\nu (t+1)} = \lambda_k^{\omega (t+1)}
    \end{align}
    Here, $t$ denotes the iteration index. A consequence of this trivial recursion is that the multipliers associated with the same shared constraint are equal at the solution. Therefore, if ALGAMES converges, it converges to a Normalized Nash Equilibrium (NNE) in the sense of Rosen \cite{Rosen1965}. An NNE is a GNE with the additional requirement that the multipliers associated with shared constraints are equal. This reasoning was applied by Dreves to a potential reduction method \cite{Dreves2011}. We transcribe it in the augmented Lagrangian context. At an NNE, because the multipliers are equal, the price to pay for violating the collision-avoidance constraint is the same for both players. This can be interpreted as enforcing a notion of ``fairness'' between the players in addition to optimality. One interesting characteristic of NNE, compared to GNE, is that they are not subject to the nullspace issue described in Section \ref{sec:fat_matrix}. Indeed, thanks to the additional constraints enforcing equality between multipliers, active constraints no longer introduce more columns than rows in the KKT system. 

\section{MPC Implementation of ALGAMES} \label{sec:mpc}
    In this section, we propose an MPC implementation of the algorithm that provides us with a feedback policy instead of an open-loop strategy and demonstrates real-time performance. We compare this MPC to a non-game-theoretic baseline on a crowded ramp merging which is known to be conducive to the ``frozen robot" problem. 
    
    \subsection{MPC Feedback Policy}
    The strategies identified by ALGAMES are open-loop Nash equilibrium strategies. They are sequences of control inputs. On the contrary, DDP-based approaches like iLQGames solve for feedback Nash equilibrium strategies that provide a sequence of control gains.
    In the MPC setting, we can obtain a feedback policy with ALGAMES by updating the strategy as fast as possible and only executing the beginning of the strategy. This assumes a fast update rate of the solution. To support the feasibility of the approach, we implemented an MPC on the ramp merging scenario described in Figure \ref{fig:ramp_merging & intersection}.a. There are 3 players constantly maintaining a 40 time step strategy with 3 seconds of horizon. We simulate 3 seconds of operation of the MPC by constantly updating the strategies and propagating noisy unicycle dynamics for each vehicle. We compile the results from 100 MPC trajectories in Table \ref{fig:benchmark & monte_carlo & mpc_table}.c. We obtain a $69$ Hz update frequency for the planner on average. We observe similar performance on the intersection problem defined in Figure \ref{fig:ramp_merging & intersection}.b, with an update frequency of $66$ Hz.
    
    
    \subsection{``Unfreezing'' the Robot}
\begin{figure}
    \includegraphics[width=8.3cm, height=3.5cm]{tikz_figures/mpc_rank_comparison.tikz}\hfill
    \footnotesize{} \textbf{(a)} Monte Carlo analysis with 100 randomly sampled initial states, we record the position of the merging vehicle in the traffic at the end of the $6$-second simulations.  
    \vspace{+2mm}

    \includegraphics[width=8.3cm]{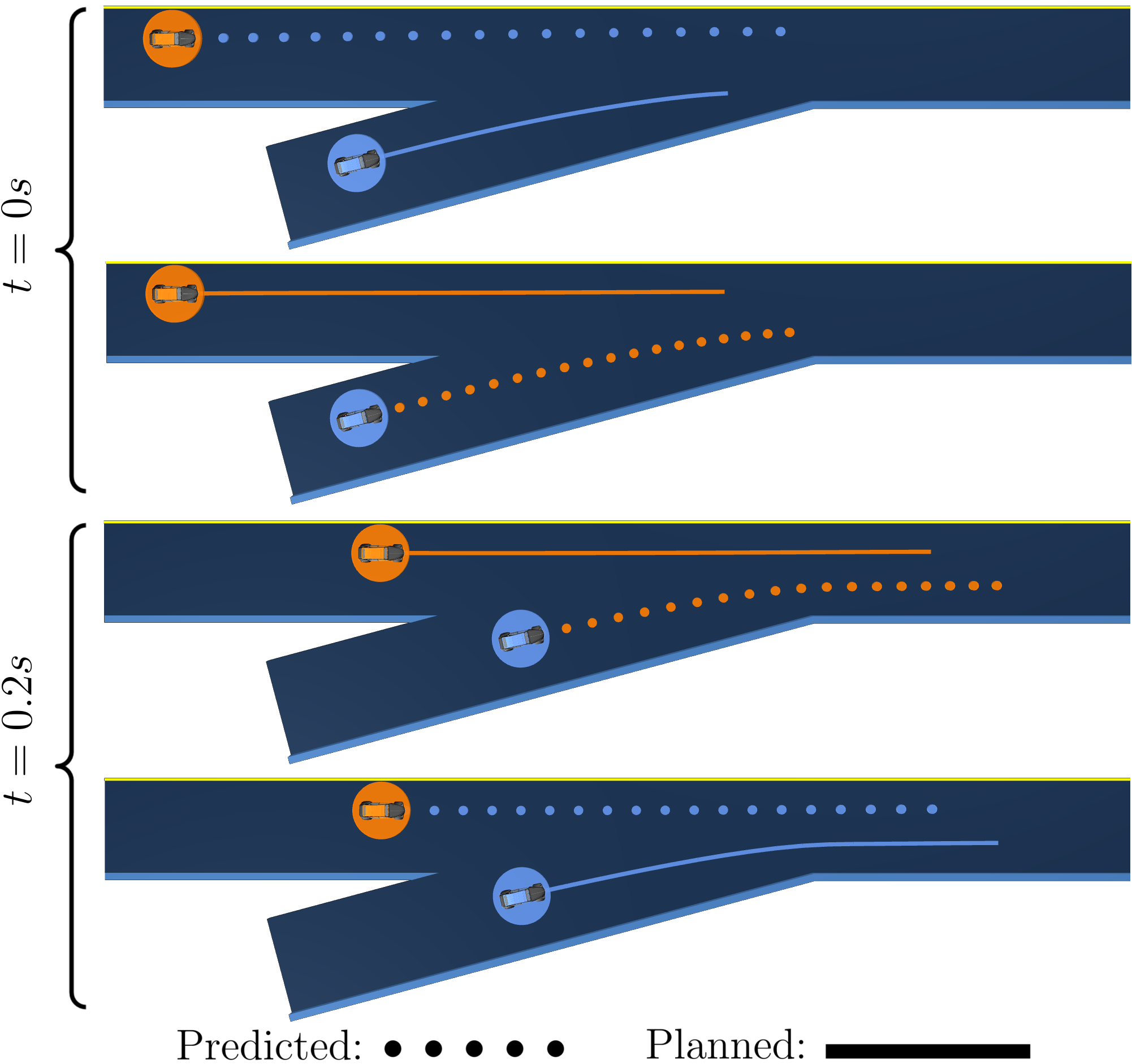}
    \textbf{(b)} Example of a Nash equilibrium mismatch. At $t=0s$, the blue car plans to merge, after letting the orange car go first. Conversely, the orange car plans to let the blue car merge first. At $t=0.2s$, the mismatch is resolved, both players converged to the same Nash equilibrium solution, where the vehicle on the ramp goes first.
    \vspace{+2mm}
    
    \includegraphics[width=8.3cm, height=3.3cm]{tikz_figure/equilibrium_mismatch.tikz}
    \textbf{(c)} Monte Carlo analysis with 100 randomly sampled initial states. Because we provided each player with a very different initial guess, we observe a mismatch between the two players for $1/3$ of the samples. I.e. the two players predict significantly different future trajectories at $t=0s$. However, the number of mismatches rapidly decreases to $2/100$, as both players converge toward the same Nash equilibrium.   

    \caption{We evaluate the ability of ALGAMES to avoid the ``frozen robot'' problem and to handle Nash equilibrium non-uniqueness.}
    \label{fig:mpc_rank_comparison & mismatch}
\end{figure}

    To illustrate the benefits of using ALGAMES in a receding-horizon loop, we compare it to a non-game-theoretic baseline MPC. With this baseline, the prediction step and the planning step are decoupled. Specifically, each agent predicts the trajectories of the surrounding vehicles by propagating straight, constant velocity trajectories. Then, each agent plans for itself assuming these predicted trajectories are immutable obstacles. 
    We test these two controllers on a challenging scenario where a vehicle has to merge on a crowded highway as presented in Figure \ref{fig:mpc_frozen_robot}. We perform a Monte Carlo analysis by uniformly sampling the initial state, $x_0$, around a nominal state with perturbations corresponding to a $\pm 2.5m$ longitudinal displacement, $\pm 25 cm$ lateral displacement,  $\pm 3^\circ$ in angular displacement for each car.
    Given the initial state, the vehicle on the ramp should be able to merge between the blue and orange cars or the orange and green cars, taking the $2^{nd}$ and $3^{rd}$ place respectively. However, waiting for all cars to pass before merging into $4^{th}$ place is not a desirable behavior. Indeed, with such a policy, the merging vehicle has to slow down significantly and could get stuck on the ramp if the highway does not clear. We run ALGAMES in a receding horizon loop and the baseline MPC to generate $6$-second trajectories for 100 different initial states. We record the position of the merging vehicle at the end of the simulation and compile the results in Figure \ref{fig:mpc_rank_comparison & mismatch}.a.
    
    We observe that the ``frozen robot'' problem occurs with the baseline MPC for $85\%$ of the simulations. An interpretation of this result is that the vehicle on the ramp cannot find a merging maneuver that is not colliding with its constant-velocity trajectory predictions. Since there is no feasible merging maneuver, the only option left is to wait for the other vehicles to pass before merging.
    
    On the contrary, by running ALGAMES in a receding-horizon loop, the vehicle merges into traffic in $2^{nd}$ or $3^{rd}$ place in $96\%$ of the simulations (Figure \ref{fig:mpc_rank_comparison & mismatch}.a). ALGAMES avoids the ``frozen robot'' pitfall in most cases by gradually adjusting its velocity to merge with minimal disruption to the traffic (Figure \ref{fig:mpc_frozen_robot}).

    \subsection{Non-Uniqueness of Nash Equilibria in Practice}
    We assess the effect of the non-uniqueness of Nash equilibria in the MPC context. We focus on the coordination issue, that players may face when there exists multiple Nash equilibria. In our experiment, each car independently runs ALGAMES as an MPC policy. Each car plans for itself and predicts the other vehicles' trajectories. We purposefully provide each player with a very different initial guess, in order to generate a mismatch between the Nash equilibrium solution that each player converges to. We simulate this on a ramp merging scenario with two players. In this scenario, an example of Nash equilibrium mismatch could be that both players think they let the oher player go first (Figure \ref{fig:mpc_rank_comparison & mismatch}.b). The results, presented in Figure \ref{fig:mpc_rank_comparison & mismatch}.c, suggest that most of the mismatches disappear rapidly after the initialization, i.e both players converges to the same Nash equilibrium. This can happen, for instance, when one Nash equilibrium is no longer feasible because it violates the bounds on the control inputs or the boundaries of the road. This positive results mitigates the concern caused by the potential occurrence of non-unique Nash equilibria. Nevertheless, it is also important to analyze the failure cases, where the two Nash equilibrium solutions found by the two players do not coincide. Typically, in these cases, each player's solution remains fairly constant and does not oscillate between multiple equilibria. In such circumstances, it would be appropriate to estimate the equilibrium that the other player is following in order to switch to this equilibrium. Peters et al. demonstrated the feasibility of this approach in similar scenarios, using a particle filter \cite{Peters2020}.
    
\section{Conclusions}
    We have introduced a new algorithm for finding constrained Nash equilibrium trajectories in multi-player dynamic games. We demonstrated the performance and robustness of the solver through a Monte Carlo analysis on complex autonomous driving scenarios including nonlinear and non-convex constraints. We have shown real-time performance for up to 4 players and implemented ALGAMES in a receding-horizon framework to give a feedback policy. 
    We empirically demonstrated the ability of ALGAMES to mitigate the ``frozen robot'' problem in comparison to a non-game-theoretic receding horizon planner.
    The results we obtained from ALGAMES are promising, as they seem to let the vehicles share the responsibility for avoiding collisions, leading to natural-looking trajectories where players are able to negotiate complex, interactive traffic scenarios that are challenging for traditional, non-game-theoretic trajectory planners. For this reason, we believe that ALGAMES could be a very efficient tool to generate trajectories in situations where the level of interaction between players is strong. Our implementation of ALGAMES is available at \texttt{\href{https://github.com/RoboticExplorationLab/ALGAMES.jl}{https://github.com/Robotic}} \\
    \texttt{\href{https://github.com/RoboticExplorationLab/ALGAMES.jl}{ExplorationLab/ALGAMES.jl}}.

\bibliographystyle{ieeetr}
\bibliography{reference}

\end{document}